\documentclass[sigconf]{acmart}

\usepackage{booktabs} 
\usepackage{subcaption}
\usepackage{multirow}
\usepackage{filecontents}
\setcopyright{rightsretained}
\acmDOI{10.475/123_4}

\acmISBN{123-4567-24-567/08/06}

\acmConference[N/A]{}{2017}{} 
\copyrightyear{2017}

\begin{document}
\title{Where's YOUR focus: Personalized Attention}

\author{Sikun LIN}
\affiliation{%
  \institution{Hong Kong University of Science and Technology}
  \state{Hong Kong} 
}
\email{sklin@connect.ust.hk}

\author{Pan Hui}
\affiliation{%
  \institution{Hong Kong University of Science and Technology}
  \state{Hong Kong} 
}
\email{panhui@cse.ust.hk}


\begin{abstract}
Human visual attention is subjective and biased according to the personal preference of the viewer, however, current works of saliency detection are general and objective, without counting the factor of the observer. This will make the attention prediction for a particular person not accurate enough. In this work, we present the novel idea of personalized attention prediction and develop Personalized Attention Network (PANet), a convolutional network that predicts saliency in images with personal preference. The model consists of two streams which share common feature extraction layers, and one stream is responsible for saliency prediction, while the other is adapted from the detection model and used to fit user preference. We automatically collect user preference from their albums and leaves them freedom to define what and how many categories their preferences are divided into. To train PANet, we dynamically generate ground truth saliency maps upon existing detection labels and saliency labels, and the generation parameters are based upon our collected datasets consists of 1k images. We evaluate the model with saliency prediction metrics and test the trained model on different preference vectors. The results have shown that our system is much better than general models in personalized saliency prediction and is efficient to use for different preferences.

\end{abstract}

%
%

\ccsdesc[500]{Computing methodologies~Interest point and salient region detections}
\ccsdesc[300]{Human-centered computing~Human computer interaction (HCI)}

\keywords{Saliency Prediction, Personalized HCI, Deep Network}

\maketitle

\section{Introduction}
\label{sec:intro}
Attention is a personalized experience, different people may focus on different regions even they are facing at a same scene: in a park containing both entertainment equipment and children playing around, attention of parents may be focused on children, whereas children may focus on the equipment. Predicting correctly where the attention is for each user is crucial for a Human-Computer-Interaction (HCI) application. With recent advances in deep learning and the improvement of computation powers, vision tasks such as object detection and saliency prediction achieve higher accuracy and faster implementations. It is feasible now to create a model for predicting user attention that can fit user preferences differently. \textit{Preference} here means the different tendencies to focus on various objects, which is inherently consistent for one particular user. 

Such a model can be a part of an augmented reality (AR) recommendation system that retrieves and displays only information that the user is in favour of, focusing the post processing only on useful regions. It is also helpful in achieving customized scene description or video summarization for individual users. The state-of-the-art saliency prediction works are all trained and evaluated on objective datasets \cite{jiang2015salicon, mit-saliency-benchmark, cheng2015global}. To fulfill the increasing demand of providing more fitting user experience, we propose in this work, Personalized Attention Network (PANet), a deep architecture that can predict personalized saliency areas.

PANet consists of two streams of convolutional neural networks (CNNs) that share common feature extraction layers. The model takes three inputs: raw image to be processed, user-defined detailed class to super category mapping, and user preference vector on super categories. Given the input image, PANet will extract its deep features at multiple scales and pass them to two streams: the saliency prediction stream will generate a saliency map without the influence of user preference, and the preference fitting stream will utilize object detection model architecture to generate a preference map according to the input preference. After combining the results gotten from two streams, post-processing including adding a centrer prior will be done and the prediction result will be given as a pixel-level saliency map that fits this particular user. To train a PANet model, we regress pixel-wisely on the ground truth, which is generated dynamically in the training generator given the input preference.

\section{Related Works}
\label{sec:relatedwork}
Although state-of-the-art saliency prediction models perform well in terms of evaluation metrics, they are all measured on benchmark datasets considering no personal preference factor. On the other hand, a subjective version of saliency prediction is needed in the real world for HCI applications. In our work, we combine object detection and saliency prediction techniques to build a model for predicting personalized attention. 

\subsection{DNNs for object detection}
Traditional detection pipeline typically consists of multiple steps: feature \cite{lowe1999object, bay2006surf, rosten2010faster} detection and descriptor extraction \cite{calonder2010brief, alahi2012freak} , feature representation \cite{csurka2004visual,perronnin2010large}, and classifiers \cite{burges1998tutorial, viola2001robust} applied in a sliding window manner \cite{felzenszwalb2008discriminatively} or in selected image sub-regions \cite{uijlings2013selective}. After CNN gains its popularity, \cite{szegedy2013deep} tried to solve detection problem in a regression manner, however it only achieves 30.5\% mAP on PASCAL VOC2007 \cite{pascal-voc-2007}. As CNN is good at classification tasks, R-CNN \cite{girshick2014rcnn} tries to combine region proposal and CNN-based classification for detection, whose good performance makes region proposal deep detection systems popular. SPP-net \cite{he2014sppnet} solves the issue of repeated calculation and speed up the whole process by putting the feature extraction convolutional layers at the beginning and introducing spatial-pyramid pooling layer, making later classification layers share the features. Fast R-CNN \cite{girshick2015fast} then refines ROI pooling layer that generates potential bounding boxes, further making the whole model end-to-end trainable. Faster R-CNN \cite{ren2015faster} tackles the speed bottleneck, selective search, by proposing Region Proposal Network (RPN), letting the network itself learn to generate region proposals. 

Even faster implementations are achieved by omitting region proposals completely and segment image into default grid cells and anchor boxes, modeling detection as regression at each cell or anchor box, 
and simultaneously learn multiple targets: object class, bounding box position, and prediction confidence. 
YOLO \cite{redmon2016yolo} segments input images into \(7\times 7\) grid cells, each responsible for predicting 2 bounding boxes and one object class.This results in a faster speed, but at the same time imposes the limitation to detect small or overlapped objects. SSD \cite{liu2016ssd} performs better as it generates default anchor boxes in different aspect ratios at each cell and at multiple feature scales. Class prediction is not bound with cells but with anchor boxes, making it more flexible in predicting multiple objects within a small sub-region. It also replaces fully connected prediction layers in YOLO with convolutional ones, further increasing the speed. To solve the detection problem with small objects, YOLO v2 \cite{redmon2016yolo9000} fine-tunes the classifier network with higher input resolution to make high-resolution detection part adjust better. It also removes the fully connected layers for bounding box prediction and use anchor boxes as in SSD. Its customized network structure and carefully designed training techniques altogether make a good end performance. Part of our model utilize the network structure of SSD, serving as part of the preference fitting stream, as well as the provider of multi-scale feature maps for the saliency prediction stream.

\subsection{DNNs for Saliency Prediction}
Saliency prediction can be categorized into three general categories: bottom-up approaches based on low-level features such as color, contrastness, texture, etc. \cite{ perazzi2012saliency, yan2013hierarchical, yang2013saliency, jiang2013salient, zhu2014saliency, cheng2015global}; top-down approaches based on high-level image features such as object knowledge \cite{ goferman2012context, shen2012unified, jia2013category}; and the combination of the two \cite{Wu2014, chen2016video}. Development in deep learning boosts the performance of saliency prediction, where saliency datasets used for training and benchmark (SALICON \cite{jiang2015salicon}, MIT300 \cite{mit-saliency-benchmark}, etc.) also make important contributions.

Recent works using deep networks in saliency detection obtain good performance, as the networks can extract more robust features than handcrafted low-level features. \cite{huang2015salicon, liu2015predicting} convert the input image into different resolutions by down-sampling, and feed them into multiple CNN streams to extract multi-scale features, and then pass the combined feature to post-processing layers to get the final saliency map. \cite{kruthiventi2015deepfix} directly uses truncated VGG-16 \cite{simonyan2014very} layers for feature extraction, and add Inception modules to capture the multi-scale information. \cite{kummerer2014deep, li2016deep, kummerer2016deepgaze} use only one CNN stream, but get features in different scales through different layers, and rescale all these feature maps to a same size for further feature merging and saliency map generation. For the saliency detection part in our work, we use the deep features extracted from VGG-16 trained on ImageNet \cite{ILSVRC15}. Meanwhile, the feature maps at different scales extracted from detection layers are merged in saliency prediction stream for further processing.

Several works \cite{judd2009learning, liu2011learning, kummerer2016deepgaze} include center prior as human generally focus their attention on the center of their eyesights. The priors are added either as a Gaussian function of the pixel distance to the center, or gotten from labeled saliency datasets. Adding center prior may have a negative influence on the result depending on the task, such as predicting eye fixations on a webpage, where human may tend to look for information on the edge menus. Our work considers only general scenes, thus center prior is added and can make positive contribution to the final result.

\section{Preference Identification}

\begin{figure*}
\begin{center}
\includegraphics[width=0.9\linewidth]{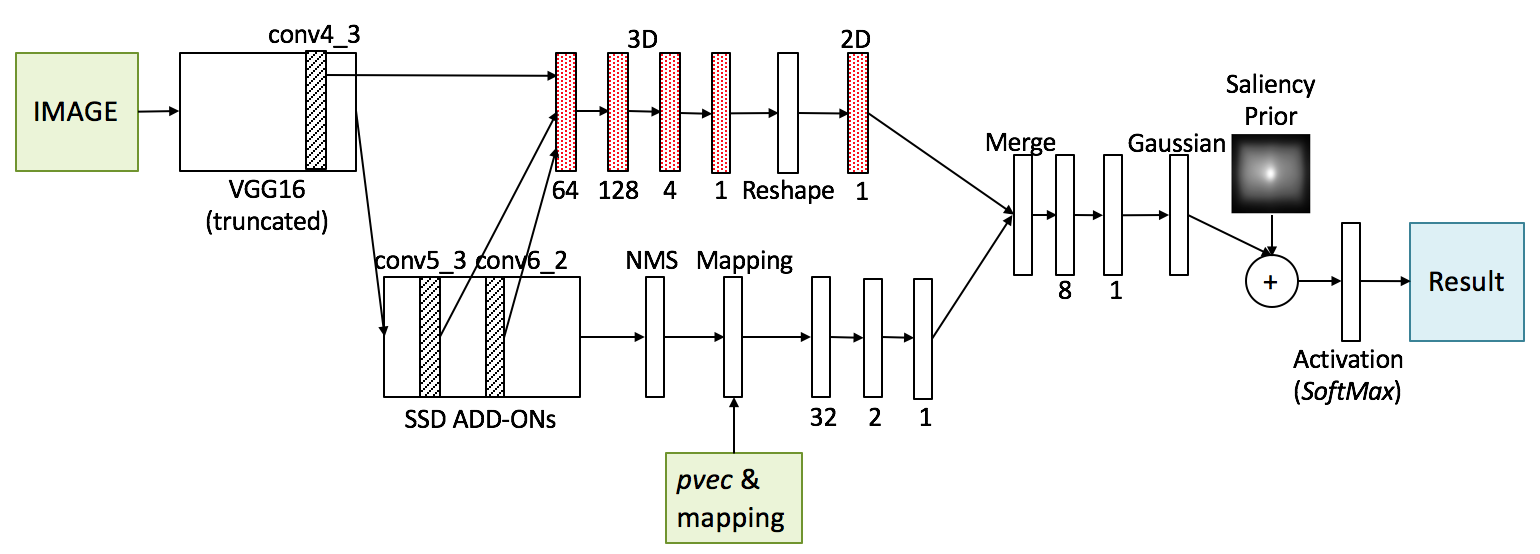}   
\end{center}
\captionsetup{justification=centering}
   \caption{PANet architecture. The model takes three inputs: original image, personal preference vector $pvec$, and detailed class to super category mapping. The upper stream is responsible for saliency detection, and the lower stream is responsible for fitting user preference. Final output is personalized saliency prediction of the input image.}
\label{fig:arch}
\end{figure*}

\label{sec:prefid}

Objects can be categorized into both detailed and super categories: for example, an apple belongs to category ``apple'' and super category ``fruit''.  
MS COCO objects have both labels, thus infers a default mapping between categories and super categories. However, this mapping has disadvantages: some super categories contain classes that are likely to co-occur, such as objects belong to ``appliance'' and ``kitchen'', and their preference scores prone to be similar to each other for a fixed user. A more meaningful mapping should merge the super categories that tend to occur in a same environment or have a similar usage. In addition, this default mapping may not fit individual needs: one user may want a mapping points to ``out-door'' and ``in-door'', whereas another user may want ``human'' and ``non-human'' super categories to represent his preference. These limitations make a general default mapping not suitable for a personalized system, so we leave the freedom of defining the mapping to user themselves.

To collect the preference of a particular user, we pass images in user's smart device into a trained detection model, in our case the SSD model. We only select image files that have last modified dates within three months to get current preference of the user. By counting the occurrence of objects falling into different categories and taking into account the prediction confidence, we can roughly know the user preference. The preference towards a particular super category $SCat_i$ is defined as: \[\mathit{Pref}_i = \sum_{x\in \mathit{SCat}_i} \mathit{Conf}_x\ ,\]
in which $x$ is the object belongs to this category, and $\mathit{Conf}_x$ is the confidence of that prediction. In this manner, more occurrence of the objects in category $SCat_i$ will lead to a higher preference of it, and detection with low confidence will not influence the preference much. After iterating over all the images, the preference vector can be obtained by normalizing the preference values to between 0 and 1: \[\mathit{Pref}_i = \frac{\mathit{Pref}_i}{\max\limits_j \mathit{Pref}_j}\ ,\]
then we can get the final preference vector of the user: \(pvec = [\mathit{Pref}_1, ..., \mathit{Pref}_n]\), where n is the total number of super categories.
This approach may suffer from accuracy problem when trained detection model covers not enough categories. Thus we offer users the chance to rate preference by themselves: we first calculate $pvec$ and present it to the user. If user thinks this $pvec$ cannot truly reflect his preference, he can then rate each super category respectively.
\section{Model Architecture}
\label{sec:arch}

As shown in Figure \ref{fig:arch}, PANet contains two streams: the upper one is for predicting the general saliency information, and the lower one is for fitting user preference upon the information gotten from detection layers. They share common image features that extracted by VGG-16 without its final classification layers as well as the convolutional layers in the detection part.

\subsection{Saliency prediction stream}
Saliency prediction part is the upper stream in Figure \ref{fig:arch}. In order to combine multi-scale features of the input image for saliency prediction, but without an extra feature extraction stream as in the \cite{huang2015salicon, li2016visual}, our model uses the features extracted in different layers from VGG-16 and SSD customized layers. Features at three different scales are used: conv4\_3, conv5\_3, conv6\_2 (layer names consistent with those in \cite{rykov8}), with size \(38\times38\), \(19\times19\), \(10\times10\) respectively. Latter two feature maps are up-sampled to the same size as the first one. In our implementation, we find that combining features from the second and third scales can notably improve saliency prediction accuracy.
After rescaling, feature maps are combined as a 3-dimensional tensor, with size \(38\times38\times3\) and 512 channels. The combined tensor is then passed through four 3-dimensional kernel 1 convolutional layers, with 64, 128, 4, 1 feature channels respectively. Then the network reshapes the tensor back to 2 dimension with size \(38\times38\) and 3 channels, and passes it through one more \(1\times1\) convolutional layer, from which the network outputs the final result with a single feature channel for the saliency detection stream.

\subsection{Preference fitting stream}
\label{sec:preffit}

This stream starts with detection layers labeled as SSD ADD-ONs in Figure \ref{fig:arch}, which is the same as those customized layers in SSD model, containing its featured anchor box generation layers and produces feature maps at multiple scales. The output of this part is a concatenation of object category, confidence and coordinate information, which needs to be converted back to a image-shaped tensor in our customized NMS and Mapping layers for further processing.

 Non-Maximum Suppression (NMS) operation is performed on above output to collapse overlapped predictions to one. In NMS layer, network also choose whether to keep a detection based on its confidence. The value of confidence threshold varies for different datasets. For PASCAL VOC2007, threshold 0.6 is enough to get most of the true positives kept, and eliminate most false positives. However, for MS COCO dataset that contains many small objects, the threshold needs to have a smaller value so as to get a satisfactory recall: we set it to 0.5, which can detect most of the small objects and have a reasonable false positive rate.

\begin{figure}
\includegraphics[height=0.8in, width=1.5in]{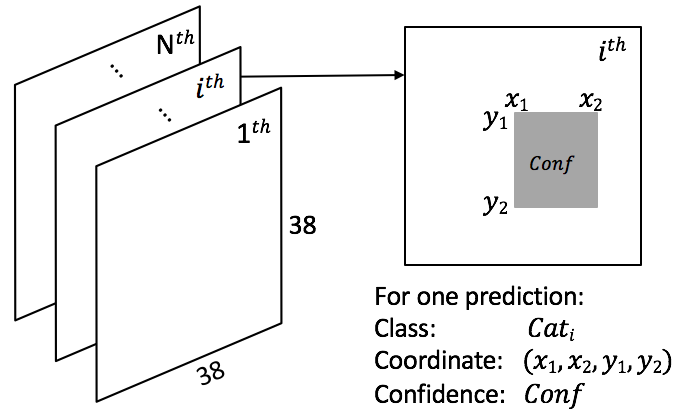}
\caption{NMS layer: Tensor creation from kept predictions.}
\label{fig:NMStensor}
\end{figure}

NMS layer then translate the information representing those predictions with high confidence into a 2-dimensional tensor in image space, as shown in Figure \ref{fig:NMStensor}. To merge with the saliency prediction stream later, the created tensor served as the output of NMS layer is set to have a size of \(38\times38\), and its channel number \(N\) is the same as the number of detailed classes. Each channel represents the prediction of one particular class. For an input image, if there exists predictions of objects in category \(Cat_i\), then the \(i^{th}\) channel of the tensor will have non-zero pixels according to the predicted position and prediction confidence. The value at each pixel \((x,y)\) is \(\max(Conf_1, \cdots, Conf_k)\), where \(Conf_1, \cdots, Conf_k\) are the confidence of predictions that have bounding boxes enclosing pixel \((x,y)\). 

\begin{figure}
\includegraphics[height=0.8in, width=1.5in]{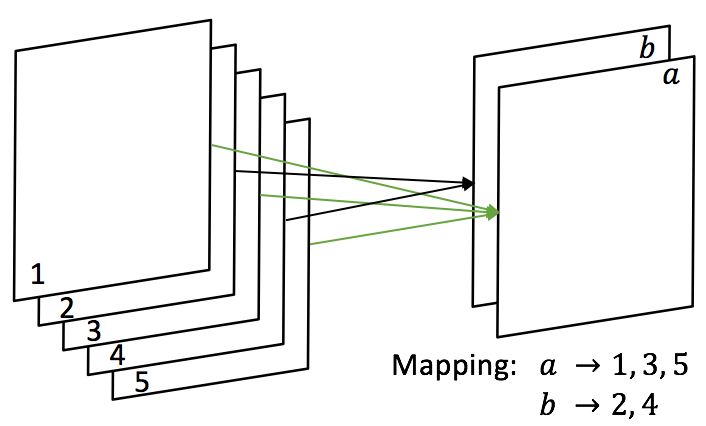}
\caption{Mapping Layer: Combine multiple channels given mapping between super categories to categories.}
\label{fig:mapping1}
\end{figure}
\begin{figure}
\includegraphics[height=0.8in, width=1.5in]{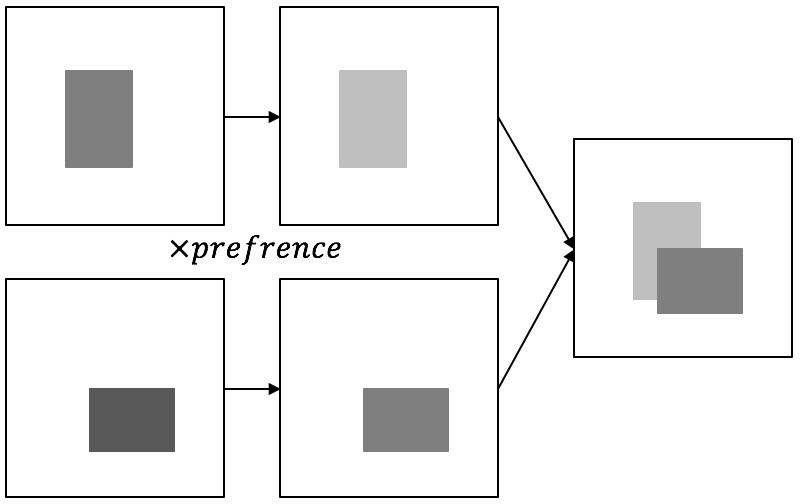}
\caption{Mapping Layer: An example of a combined channel from two channels.}
\label{fig:mapping2}
\end{figure}

The output channels of NMS layer are in detailed categories, while Mapping layer combines them into channels representing super categories. Mapping layer takes in two extra inputs: user preference vector and defined mapping between super categories to categories. Given such a mapping:
\[\begin{split}
    \{SCat_1 &\rightarrow Cat_{11}, \cdots, Cat_{1k_1};\\
    &\vdots
    \\SCat_n &\rightarrow Cat_{n1}, \cdots, Cat_{nk_n}\} \ ,
\end{split}
\]
tensor channels representing \(Cat_{ij} (\forall j)\) will be merged into a single channel \(SCat_i\): Figure \ref{fig:mapping1} shows for an particular mapping, how multiple channels representing different categories are combined to channels that represent super categories. 
The pixel-wise value of the new channel representing \(SCat_i\) is: \(SCat_i = \underset{j}{\max} (Cat_{ij}\times pvec[SCatId_{i}])\), where \(pvec[SCatId_{i}]\) represents the preference towards category \(SCat_{i}\). The process of this mapping operation is summarized as in Figure \ref{fig:mapping2}, and it is the key layer that makes the model take in and fit user preference.

\subsection{Merging two streams}
The model merges two streams together by tensor concatenation, and two \(1\times1\) convolutional layers with channel number 8 and 1 are added for more nonlinearity. Further more, as people tend to focus on the central part of their eyesight, we add a center prior to our model before the final activation layer, and we generate this prior map from saliency labels in SALICON dataset by summing up all the saliency ground truth SAL\(_{gt}\) in the dataset, and then normalize it to \([0,1]\):
\[\begin{split}
    &prior = \sum \textrm{SAL}_{gt}\ ,
    \\& prior = \frac{prior-\min(prior)}{\max(prior)-\min(prior)}
\end{split}
\]
Finally, a softmax activation layer is added to output the final prediction as a probability map.

\section{Data}

Currently there is no personalized attention annotation available. To train our model, we need ground truth data labeled with consistent preference. In our implementation, we first collect a small dataset labeled by two subjects with their own preferences, and then we generate more ground truth data according to our collected labels.

\subsection{Data Collection}

\begin{figure}[t]
\centering
\includegraphics[width=0.35\textwidth]{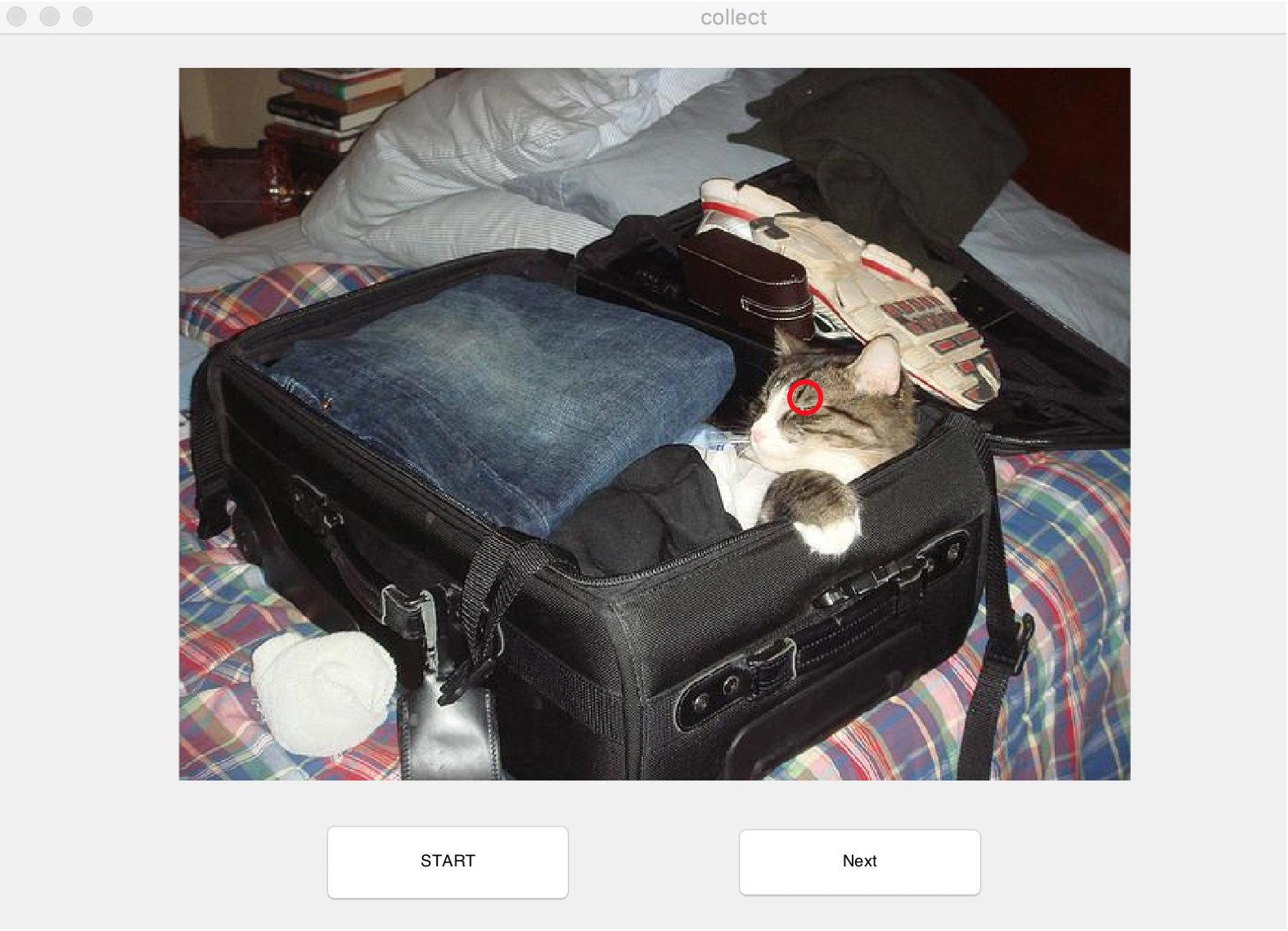}
\caption{Matlab GUI for collecting user labels.}
\label{fig:gui}
\end{figure}

To collect this dataset, we asked two subjects A and B to firstly define their preference super categories, then rate each category in the scale of 0 to 10, the higher rating means the subjects is more interested in that category. We re-scale their rating to 0 to 1 to get their corresponding $pvec$, listed in Table \ref{table:collect_pvec}. After this, we select a subset of images from SALICON dataset which consists of 15k images. The selected images contain at least one object that belongs to the categories rated larger than 5, as labeling images having both preferred objects and other objects can make the label more useful. This turned out to be 3126 images for subject A and images 3974 for subject B: we randomly choose 500 images for each subjects to label. 
\begin{table}[h]
\centering
\begin{tabular}{ |c| c| }
 \hline
 A & $[cat, car, others] = [1.0, 0.8, 0.2]$\\ 
 \hline
 B & $[keyboard, food, person, others] = [1.0, 0.8, 0.5, 0.3]$ \\
 \hline
\end{tabular}
\caption{Subject preference vectors in our collected dataset.}
\label{table:collect_pvec}
\end{table}

\begin{figure*}
\centering
\begin{subfigure}[t]{0.46\textwidth}
\includegraphics[width=\textwidth]{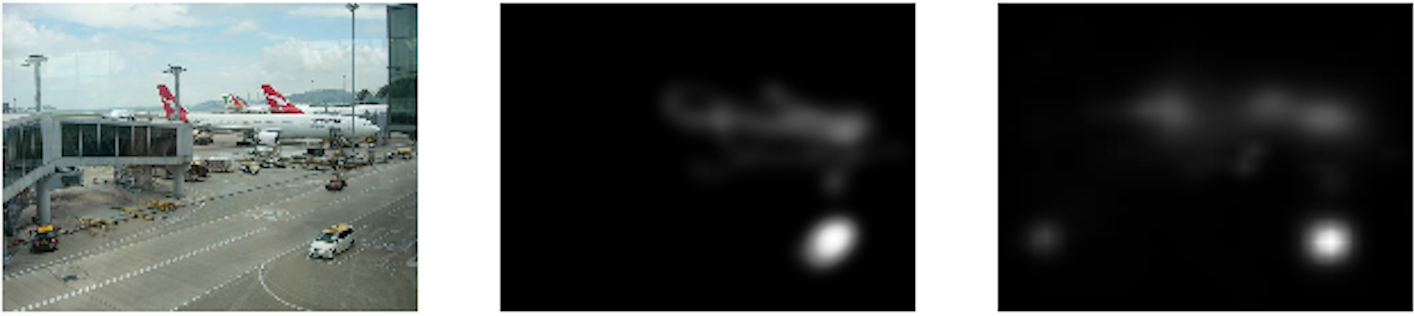}
\end{subfigure}\hspace{0.03\textwidth}
\begin{subfigure}[t]{0.46\textwidth}
\includegraphics[width=\textwidth]{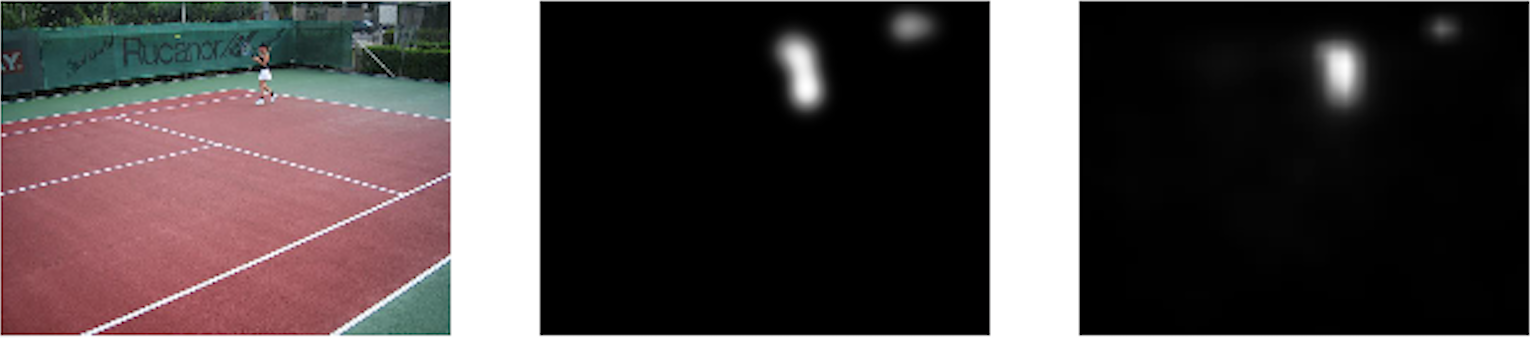}
\end{subfigure}\\[2ex]
\begin{subfigure}[t]{0.46\textwidth}
\includegraphics[width=\textwidth]{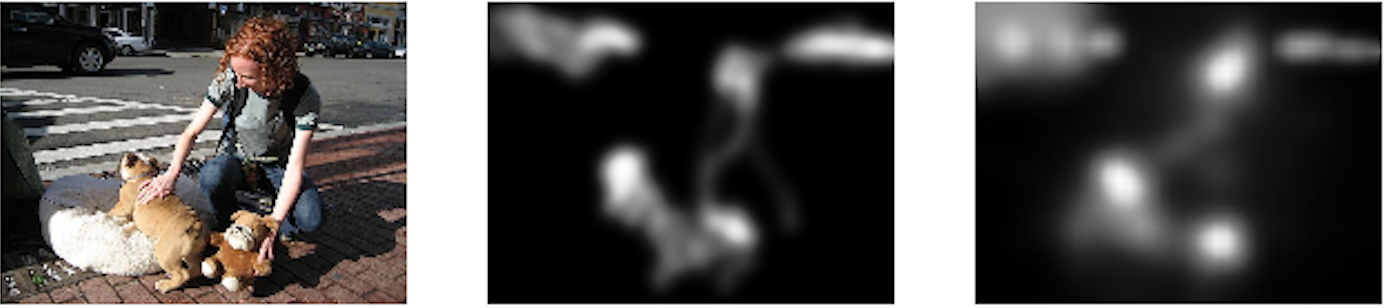}
\end{subfigure}\hspace{0.03\textwidth}
\begin{subfigure}[t]{0.46\textwidth}
\includegraphics[width=\textwidth]{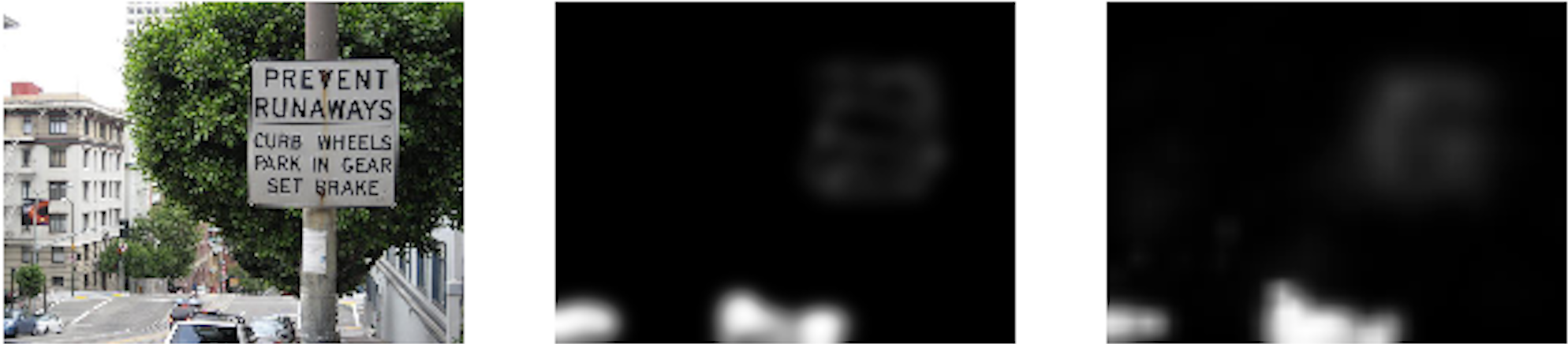}
\end{subfigure}
\captionsetup{justification=centering}
  \caption{COCO image, visualized ground truth label (subject A), our generated ground truth.}
\label{fig:gen_compare}
\end{figure*}

The presentation of image is done in Matlab R2015b, and has a GUI as shown in figure \ref{fig:gui}, in which the mouse cursor is represented as a red circle. Label collection is done through mouse tracking and has a similar procedure as \cite{jiang2015salicon}. Each image is presented to the subject and the subject is asked to move the mouse cursor to where they want to look at. A mouse click will trigger the recording process as well as finishing current recording. When the subject finishes current image annotation, he will need to hit \textit{Next} button for the next image to show up. There will be 2 seconds waiting interval showing a blank figure between two images. As in \cite{jiang2015salicon}, 50 images form a block, and subjects are allowed to take a short break between blocks. To make our result stable, we ask each subject annotation their 500 images five times, in five separate days. The final fixation annotation for each image is averaged from those five annotations. The middle column of Figure \ref{fig:gen_compare} shows the visualized results.

\subsection{Data Generation}

\label{sec:datagen}
Taking into account that preference is unique, and it is impractical to build a dataset with preferences cover all typical users, it is better to use ground truth generated dynamically based on input preference. This can make us train PANet with more data, and make the training flexible on new preferences, so that we do not need to ask each new user using our model to label thousands of images with their preferences. We generate our training ground truth data by utilizing currently available detection and fixation annotations, in our case the MS COCO dataset \cite{lin2014microsoft} and its attention annotation dataset SALICON \cite{jiang2015salicon}.
\begin{figure}
\centering
\begin{subfigure}[t]{0.23\textwidth}
\includegraphics[width=\textwidth]{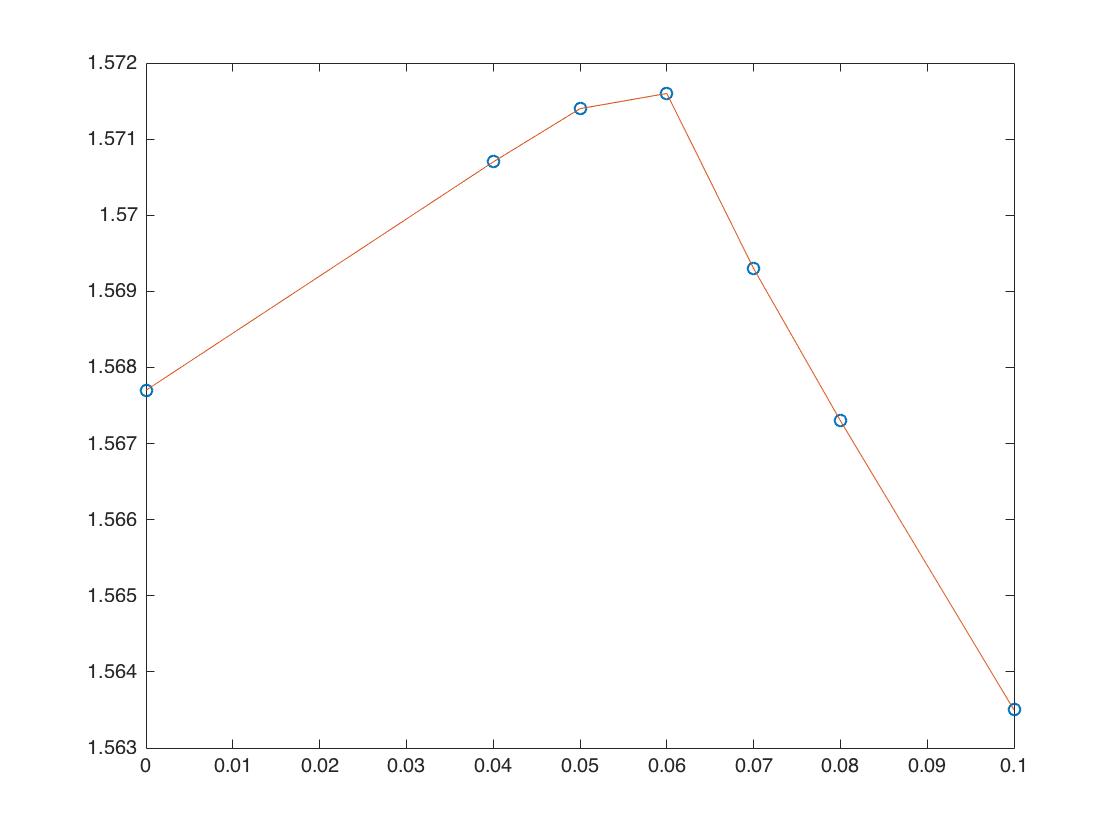}
\caption{Sum of CC and SIM at different $\alpha$, with fixed $\beta:\gamma = 0.8:0.2$}
\label{fig:gen1}
\end{subfigure}
\begin{subfigure}[t]{0.23\textwidth}
\includegraphics[width=\textwidth]{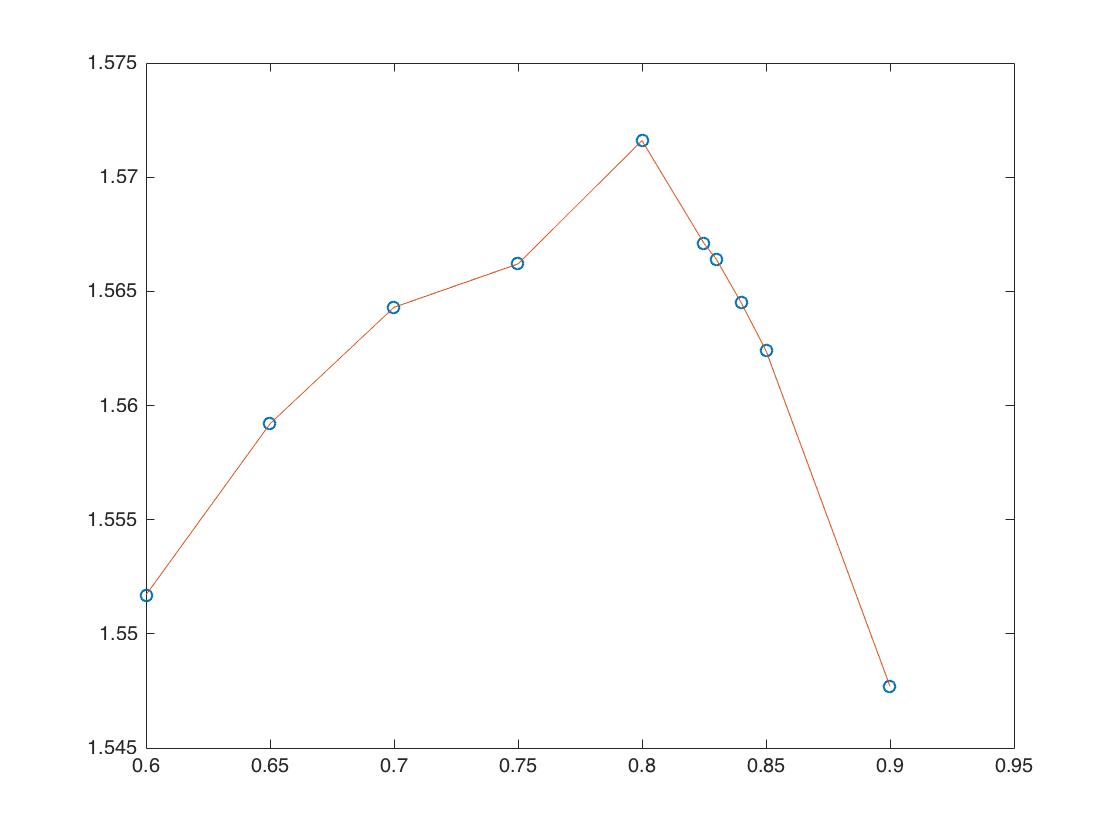}
\caption{Sum of CC and SIM at different $\beta:\gamma$, with fixed $\alpha = 0.06$. $x$ coordinates is the $beta$ ratio.}
\label{fig:gen2}
\end{subfigure}
\label{fig:gen_weights}
\captionsetup{justification=centering}
    \caption{Choosing weights for ground truth generation.}
\end{figure}

Given a particular $pvec$, we first generate a preference map $pMap$ using the ground truth bounding box positions and object category information:
\(
    pMap[x,y] = \max(pvec[SCatId[x,y]])\ ,
\)
where $SCatId[x,y]$ are the super category IDs of the objects cover pixel $[x,y]$, and we select the maximum preference when multiple objects co-occur at this position. Then we use the saliency ground truth \(\textrm{SAL}_{gt}\) in SALICON dataset as prior, redistributing the attention according to $pMap$. In addition, \(\textrm{SAL}_{gt}\) is added to make sure salient area receives fair amount of attention even it is not preferred by the user. There are also regions containing objects in the preferred categories, but have low saliency probability in SALICON annotations: to give a chance to these areas, $pMap$ itself is also added:
\[\textrm{PSAL}_{gt}=\alpha\,\textrm{SAL}_{gt} +\beta\, \textrm{SAL}_{gt} \cdot pMap+\gamma\,pMap\ ,\]
where PSAL\(_{gt}\) is the generated personal saliency ground truth with the preference $pvec$. To choose values of the weights, we first fix $\beta:\gamma$ ratio to be $0.8:0.2$, and find $\alpha=0.06$ will give the best performance, as shown in Figure \ref{fig:gen1}, in terms of average CC and SIM scores (metrics explained in Section \ref{sec:quan}). After determining the most fitting $\alpha$, we change the ratio between $\beta$ and $\gamma$, the generation performance is shown in Figure \ref{fig:gen2}, with $\beta:\gamma = 0.8:0.2$ getting the best score. Under the constraint  $\alpha+\beta+\gamma=1$, the final wights are 0.06, 0.752 and 0.188 respectively.

Generated ground truth are then normalized and saved as a probability distribution by going through a softmax:
\[\begin{split}
    &\textrm{PSAL}_{gt} = \frac{\textrm{PSAL}_{gt} - \min(\textrm{PSAL}_{gt})}{\max(\textrm{PSAL}_{gt})-\min(\textrm{PSAL}_{gt})}\ ,\\
    &x_i = \frac{e^{x_i}}{\sum_j e^{x_j}}\ ,
\end{split}\]
where $x_i$ denotes for every pixel value in each generated $\textrm{PSAL}_{gt}$. We show the comparison of labeled ground truth and our generated one in Figure \ref{fig:gen_compare}.

\section{Training}
\label{sec:training}

We train PANet in three phases. Input images fed to the whole model are rescaled to \(300\times300\), and augmented through crop, flip, and color jitter, each with probability 0.5. For all training phases, we keep a track of the minimum validation loss, and if the validation loss of three continuous epochs are all larger than the minimum one, we assume the model starts to overfit and stop training.

The first phase is pretraining a SSD model. PANet will use the weights of SSD300 layers. To get the weights trained on our desired data, we pretrain a SSD300 model on MS COCO dateset which contains 80k training images and 40k validation images. Its feature extraction layers, namely VGG-16 without the final classification layers, have the weights pretrained on ImageNet. The training process is adapted from \cite{rykov8,liu2016ssd}. According to MS COCO section in \cite{liu2016ssd}, the scale of the default boxes are set to be smaller than normal to fit small objects in this dataset. As for the default box aspect ratio, we don't use the generation method as in \cite{liu2016ssd}, instead, we first get a prior box distribution from ground truth bounding boxes, and then use it for the default box generation. We train the model for 180k iterations with batch size 20, and the learning rate starting from $3\times10^{-4}$, decreasing every epoch with a rate of 0.9.

The second phase is pretraining saliency layers. Without merging the result from the preference fitting stream, the weights of those layers responsible for saliency prediction (labeled red in Figure \ref{fig:arch}) are pretrained. This will improve model performance compared with directly training the entire model. We import and freeze the weights from previous phase into this phase in order to extract image features. The model is trained on the original SALICON dataset, which contains 10k training images and 5k validation images. The ground truth label is unbiased saliency annotation. We directly regress the predicted saliency in probability distribution to the ground truth distribution pixel-wisely. Training objective for this phase is KL-divergence:
\[D_{KL}(p|q)=\sum_i p_i \log\frac{p_i}{q_i}
\]
which is a common metric for saliency prediction models. We trained this part for 30k iterations with batch size 20, and the learning rate starts from $10^{-3}$, decreasing by 0.9 per epoch.

The final phase is training entire PANet on our dynamically generated ground truth containing 10k training images and 5k validation images. For our training, randomly generated preference vector and default MS COCO category mapping are served as extra inputs to the model.
We load and freeze the weights of VGG part and SSD add-on layers, while the weights learned for saliency prediction layers are loaded and fine-tuned. Layers without pretrained weights are initialized with random weights. The training objective is again KL-divergence, and we train the model for 50k iterations with batch size 20 and the same learning rate as in the previous phase.


\section{Evaluation}
\label{sec:evaluation}

In this section, we will first describe our results in the quantitative manner, and then show its prediction results qualitatively. As we do not want to retrain our model every time when it is applied to a new preference, we will present the experimental results of using a trained model to a new preference vector and user-defined mapping in Section \ref{sec:transfer}. We will also compare our model with a objective saliency model.

\begin{figure*}
\centering
\begin{subfigure}[t]{0.46\textwidth}
\includegraphics[width=\textwidth]{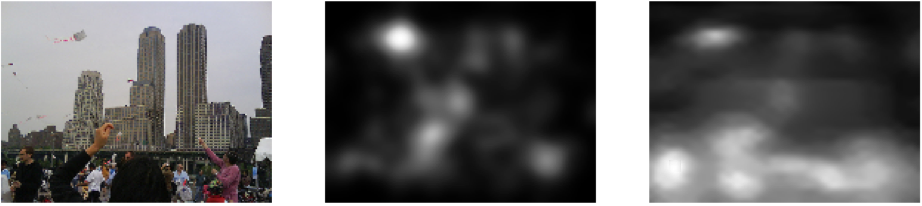}
\caption{Attention paid more to people and vehicles.}
\end{subfigure}\hspace{0.03\textwidth}
\begin{subfigure}[t]{0.46\textwidth}
\includegraphics[width=\textwidth]{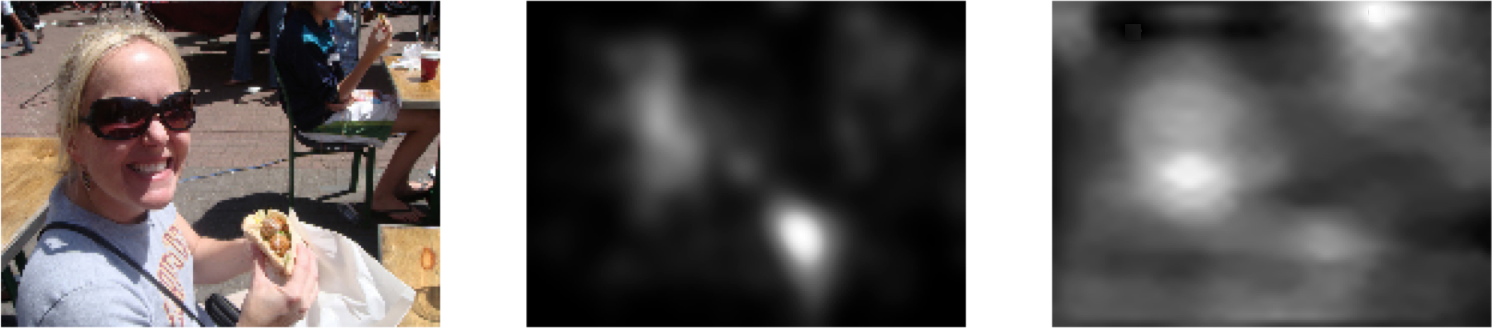}
\caption{Attention paid more to people than food, even on the back.}
\end{subfigure}
\begin{subfigure}[t]{0.46\textwidth}
\includegraphics[width=\textwidth]{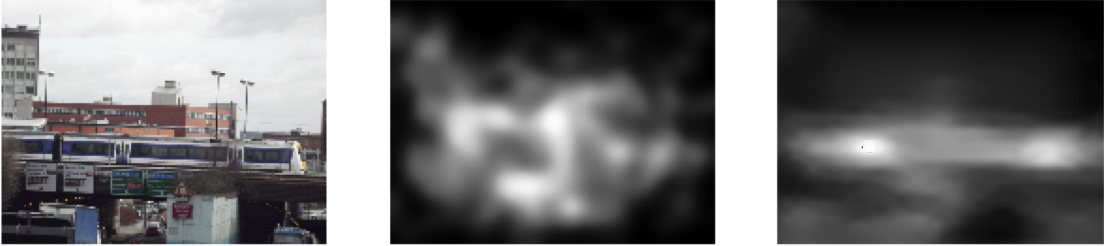}
\caption{Limit attention area to vehicle.}
\end{subfigure}\hspace{0.03\textwidth}
\begin{subfigure}[t]{0.46\textwidth}
\includegraphics[width=\textwidth]{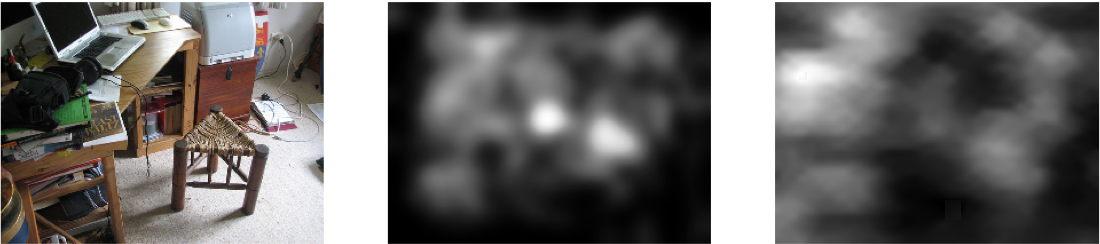}
\caption{Attention shifts from furniture to electronic devices.}
\end{subfigure}
\begin{subfigure}[t]{0.46\textwidth}
\includegraphics[width=\textwidth]{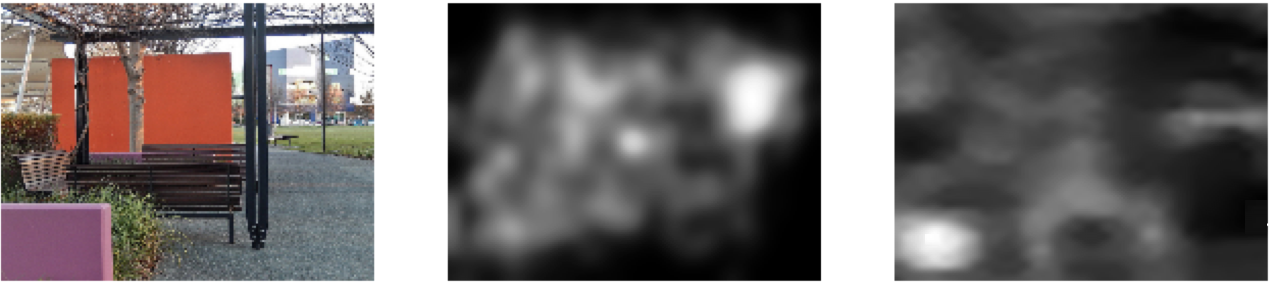}
\caption{Problematic prediction 1.}
\label{fig:w1}
\end{subfigure}\hspace{0.03\textwidth}
\begin{subfigure}[t]{0.46\textwidth}
\includegraphics[width=\textwidth]{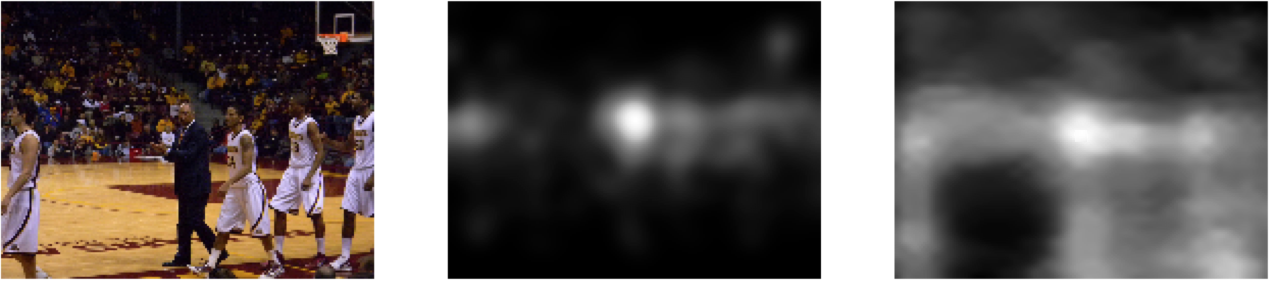}
\caption{Problematic prediction 2.}
\label{fig:w2}
\end{subfigure}
\captionsetup{justification=centering}
  \caption{COCO image, saliency ground truth, model prediction with preference vector: \\ $[$\textit{outdoor, food, indoor, appliance, sports, person, animal, vehicle, furniture, accessory, electronic, kitchen}$]$\\ = [0.833, 0.346, 0.189, 0.098, 0.934, 0.679, 0.481, 0.875, 0.081, 0.579, 0.901, 0.223].}
\label{fig:result1}
\end{figure*}
\begin{figure*}
\centering
\begin{subfigure}[t]{0.48\textwidth}
\includegraphics[width=\textwidth]{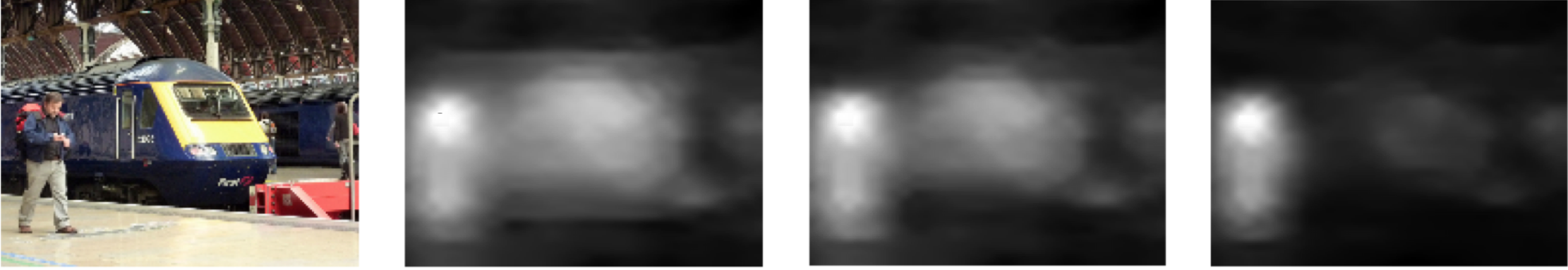}
\caption{}
\label{fig:tr1}
\end{subfigure}\hspace{0.02\textwidth}
\begin{subfigure}[t]{0.48\textwidth}
\includegraphics[width=\textwidth]{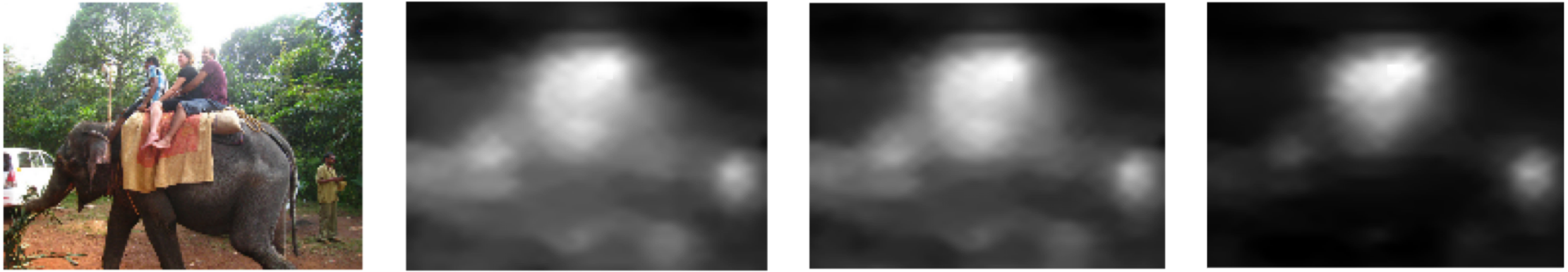}
\caption{}
\label{fig:tr2}
\end{subfigure}
\captionsetup{justification=centering}
  \caption{Effect of fine-tuning. Images are: COCO image, prediction for $pvec_1$ with original model, prediction for $pvec_2$ with original model, prediction for $pvec_2$ with fine-tuned model.}
\label{fig:trans_finetune0}
\end{figure*}

\subsection{Quantitative Results}
\label{sec:quan}

MS COCO images, thus its subset SALICON, generally contains multiple objects and covers different super categories in one image, making it suitable for our experiment. We randomly re-split the 15K images in SALICON into 7K training images, 3k validation images and 5k test images, and then train the model with preference vector shown in Figure \ref{fig:result1}. We compare our model with two baselines. The first is center prior, which sets the saliency baseline. The second is detection baseline: for each image, we highlight detected objects with detection confidence $\times$ object preference. One problem of the second baseline is that there exist images that no object is detected with confidence higher than the 0.5. This will result in zero matrices which cannot be used in metrics. To solve this, we replace these zero matrices with randomly generated matrices that are normalized to have a pixel-sum 1. In the following, we use p,q to represent predicted saliency map and dynamically generated ground truth respectively.

We calculate Pearson's Correlation Coefficient:
\[\text{CC}(p,q) = \frac{\sigma(p,q)}{\sigma(p)\times\sigma(q)} \,
\]
where $\sigma(p, q)$ is the covariance of p and q. The average CC score is 0.725. The center prior baseline has an average CC 0.420, and detection baseline has a average CC 0.493. These indicate that PANet predicts much better than the two baselines. 

Apart from the linear relationship, we also measure the $Similarity$ between predicted saliency map and our ground truth:
\[\text{SIM}(p,q) = \sum_i \min(p,q)\ ,\]
where \(\sum_i p_i = \sum_i q_i = 1\).
Two same saliency maps will have SIM$(p,q)=1$. Different from CC that is calculated symmetrically and measures false positive and false negative equally, SIM metric penalizes false positives less than false negatives, making the center prior baseline scores higher under this metric. The average score on the test set is 0.742 and the center prior baseline has SIM value 0.622 comparing with the ground truth, while the detection baseline scores 0.587.

 Kullback-Leibler divergence measures the difference between two saliency maps that are viewed as distributions, and we used KL-Judd variant in our measurement:
\[KL(p,q) = \sum_i q_i \log(\epsilon+\frac{q_i}{\epsilon+p_i})\ ,
\]
where $\epsilon = 2.2204e-16$ is a regularization constant. KL-divergence is non-symmetric, and penalizes the pixel predictions where \(q_i\) are much larger than \(p_i\). For our experiment on MS COCO with this particular preference vector, the average KLD is 0.159, and the center prior baseline is 0.316. Detection baseline has a much higher averge KLD score 11.222 as its distribution is much different from a saliency map. We will also see in Section \ref{sec:transfer} that the variance of preference vector will change KLD score significantly, making it less suitable for measuring our model performance.

Earth Mover's Distance (EMD) measures the spatial distance between distributions by computing the minimum cost of moving densities from one distribution into another. The version we use in our experiment is its linear variant as in \cite{mit-saliency-benchmark}:
\[EMD(p,q) = \min_{\{f_{ij}\}}\sum_{i,j}f_{ij}d_{ij} + |\sum_i p_i - \sum_j q_j|\max_{ij}d_{ij}\ , \]
under constraints: $(1)f_{ij}\geq0$, $ (2)\sum_j f_{ij} \leq p_i$, $(3) \sum_i f_{ij}\leq q_j$, \((4)\sum_{i,j}f_{ij} = \min(\sum_i p_i, \sum_j q_j)\), 
where $d_{ij}$ is the ground distance between $i^{th}$ and $j^{th}$ bin of the distributions, and $f_{ij}$ is the density flow between them. The average EMD score on the test set for this particular preference vector is 0.685, with center prior scoring 3.668 and detection baseline scoring 5.116. 

\subsection{Qualitative Results}
\label{qualr}

We show our model predictions in Figure \ref{fig:result1}, which are tested upon MS COCO images with the preference vector listed in the caption. We visualize the result by reshaping the final output back to \(38\times38\) and then resizing it to the original image shape. We select the result that can represent the difference between PANet and general saliency models, which means in the same image there exist objects belong to multiple super categories, and the user's preference on these categories are largely different. From the result we can see that salient areas of an input image are different in the general saliency ground truth and in our prediction results, and our prediction area fit better for this particular input preference.

Performance bottleneck is the accuracy of detection layers. When detection layers cannot recognize an object or recognizes it wrongly, predicted saliency map either pays not enough attention to correct areas, or pays attention to wrong areas. For example, in Figure \ref{fig:w1}, the foreground and background chair belong to the same category, but model wrongly detects the foreground one as an object with high preference, thus attracts more attention. Another problem of the model is that when objects in the preferred category fill the image, the predicted saliency map will have dispersive attention, as shown in Figure \ref{fig:w2}.

\subsection{Same model, different preference}
\label{sec:transfer}


\begin{figure*}
\centering
\begin{subfigure}[t]{0.23\textwidth}
\includegraphics[width=\textwidth]{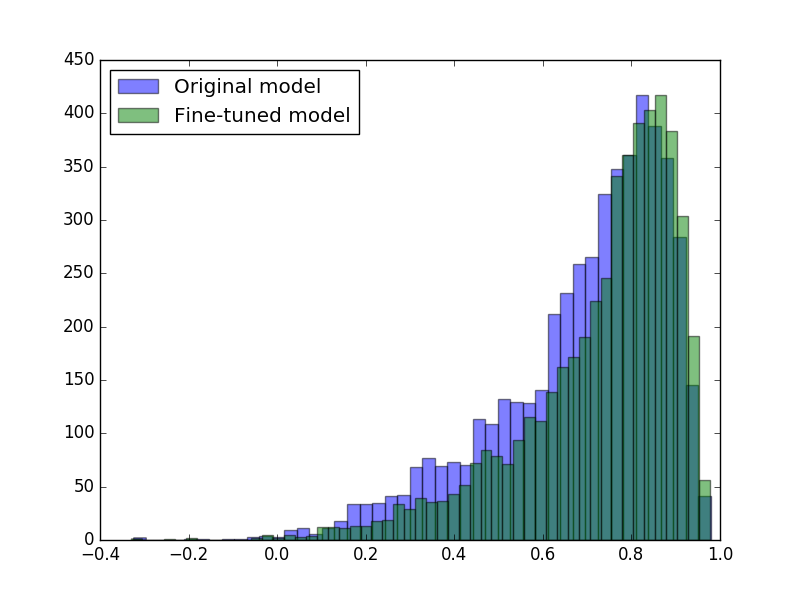}
\caption{CC}
\end{subfigure}
\begin{subfigure}[t]{0.23\textwidth}
\includegraphics[width=\textwidth]{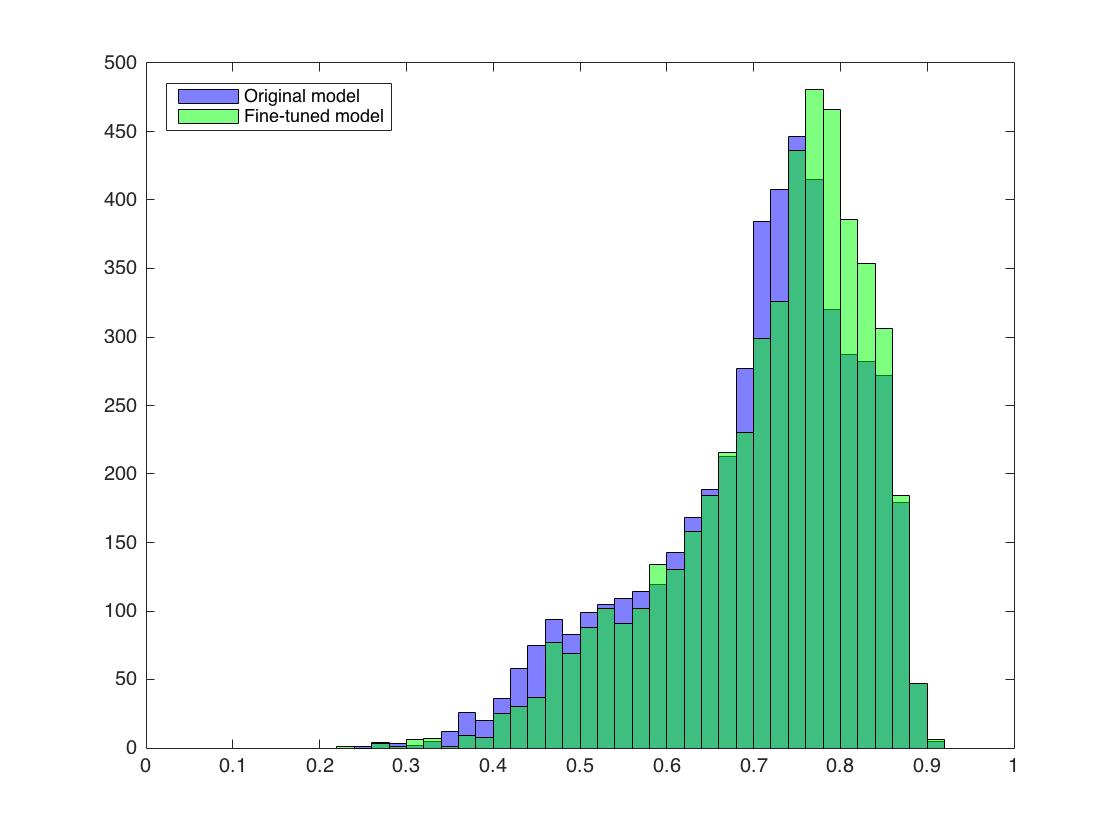}
\caption{SIM}
\end{subfigure}
\begin{subfigure}[t]{0.23\textwidth}
\includegraphics[width=\textwidth]{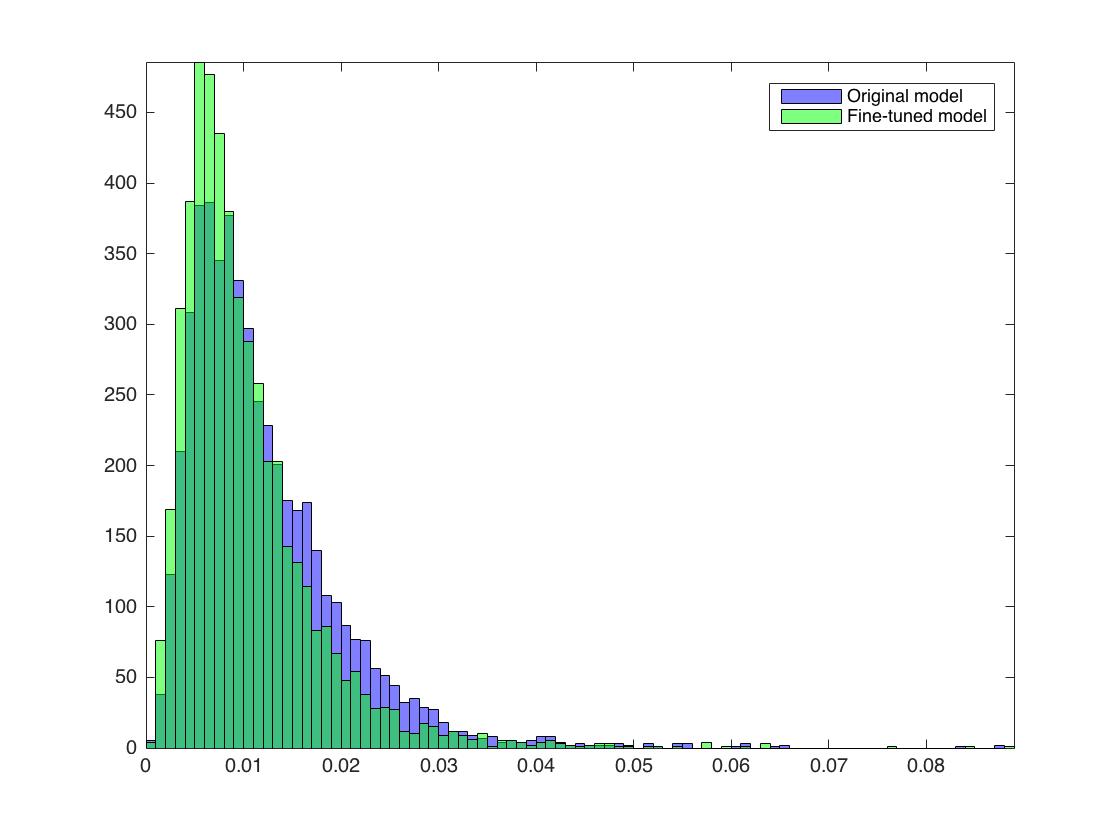}
\caption{KLD}
\end{subfigure}
\begin{subfigure}[t]{0.23\textwidth}
\includegraphics[width=\textwidth]{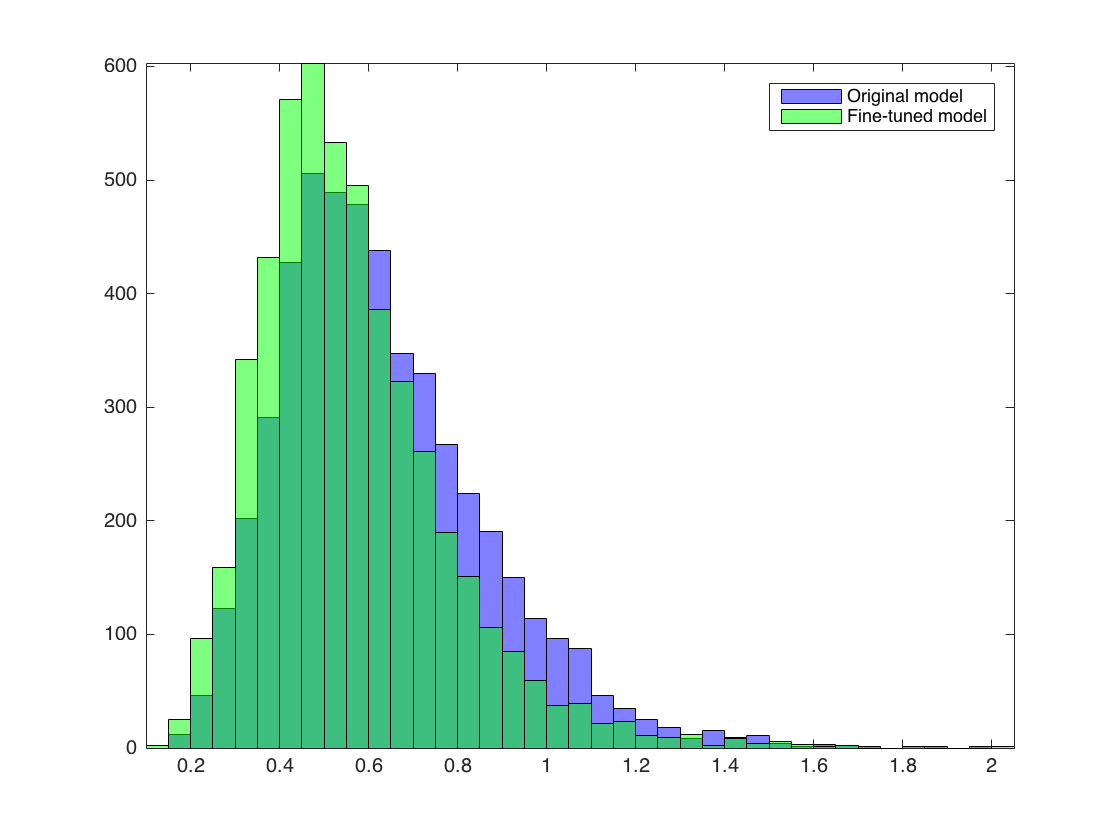}
\caption{EMD}
\end{subfigure}
\captionsetup{justification=centering}
  \caption{Performance of original model and fine-tuned model.}
\label{fig:trans_finetune}
\end{figure*}

To fit different user-defined mappings without retraining the whole model each time, we fix the output channel of the Mapping layer to 20, which sets the maximum limit on allowed number of super categories. If the number of user-defined super categories is less than 20, we automatically fill the remained channels by zero. In our experiment, we test the model with two preference vectors of different lengths on the SALICON test set (split as in Section \ref{sec:quan}): $pvec_1$ listed in Figure \ref{fig:result1} and $pvec_2$ is $[person, non-human]= [1.0, 0.05]$.
First, we directly pass $pvec_2$ and new mapping $[person\rightarrow person; \text{others}\rightarrow non\text{-}human]$ into the original model that is trained with the ground truth generated upon $pvec_1$. The results are shown in Figure \ref{fig:tr1}3 and \ref{fig:tr2}3. Comparing with Figure \ref{fig:tr1}2 and \ref{fig:tr2}2, ``person'' ares indeed get more attention, but still not good enough to reflect $pvec_2$, since $pvec_2$ is very biased. We then fine-tune the model to see the results. Loading the weights trained with $pvec_1$, and fine-tuning the model with ground truth generated by $pvec_2$ as well as changing inputs to $pvec_2$ and the new mapping, PANet converges very fast. Only three epochs (1050 iterations) are needed to make the objective function reach its lowest point, which is much less than our original training that takes 50K iterations. Figure \ref{fig:tr1}4 and \ref{fig:tr2}4 show the results predicted by the fine-tuned model, which represent $pvec_2$ much better. This indicates that given a trained model, doing saliency predictions on a new preference is easy and fast.

\begin{table}
\begin{subtable}{.5\textwidth}
\centering
\begin{tabular} { | c | c | c | c | c | c |}
\hline
 Model & $pvec$ & CC & SIM & KLD & EMD\\ 
 \hline
\multirow{2}{5em}{Original} & $pvec_1$ & 0.725 & 0.742 & 0.159 & 0.685\\
 & $pvec_2$ & 0.689 & 0.703 & 0.0120 & 0.686\\
 \hline
Fine-tuned & $pvec_2$ & 0.729 & 0.720 & 0.0102 & 0.591\\
 \hline
\end{tabular}
\caption{Average scores.}
\end{subtable}
\begin{subtable}{.5\textwidth}
\centering
\begin{tabular} { | c | c | c | c | c | c |}
\hline
 \multicolumn{2}{|c|}{} & CC & SIM & KLD & EMD \\ 
 \hline
\multirow{2}{5em}{Center prior baseline} & $gt_1$ & 0.420 & 0.622 & 0.316 & 3.668\\
 &$gt_2$ & 0.431 & 0.610 & 0.315 & 3.635\\
 \hline
 \multirow{2}{5em}{Detection baseline} & $gt_1$ & 0.493 & 0.587 & 11.222 & 5.116\\
 &$gt_2$ & 0.499 & 0.579 & 11.086 & 5.164 \\
 \hline
\end{tabular}
\caption{Baseline scores.}
\end{subtable}
\caption{Performance of original model and fine-tuned model on different preference vectors.}
\label{table:trans}
\end{table}

We also measure the performance quantitatively
as shown in Table \ref{table:trans}, and the score distributions are shown in Figure \ref{fig:trans_finetune}. When trying to predict the saliency map for a new preference vector, a fast fine-tuning will increase the performance significantly than directly using the original model. The result also demonstrates that for different preferences, model performances are different on a same metric, which might be influenced by the variance of preference scores. KL-divergence score is largely determined by $pvec$. An even preference, namely each super category receives similar attention, will likely result in a dispersive saliency map that has smaller values at each pixel position, and are more likely to be penalized by KL-divergence measure.

\subsection{Comparison with unbiased models}
To see the effect of shifting attention according to personal preference, we compare our model with a state-of-the-art general saliency prediction model $Salicon$ \cite{huang2015salicon}, testing upon the ground truth generated with $pvec_1$ and $pvec_2$ as in Section \ref{sec:transfer}. We use their codes \cite{christopherleethomas2016} to generate fixation maps of the images without any personal tendency. The results are shown in table \ref{table:unbiased}.

Comparing the predictions of our PANet and generalized models such as Salicon model, PANet performs much better in fitting user preference. The saliency maps predicted by a general model such as Salicon score similarly to the center prior. For different preferences, the performances will be different but stay in the same level. A more biased preference will lead to a slightly worse performance when using the general saliency model.

\begin{table}[t]
\centering
\begin{tabular} { | c | c | c | c | c |}
\hline
 $pvec$&CC&SIM&KLD&EMD\\ 
 \hline
$pvec_1$&0.527&0.621 &0.332 & 4.605\\
 \hline
$pvec_2$&0.525&0.594 &0.337 & 4.656\\
 \hline
\end{tabular}
\caption{Salicon model tested on the ground truth generated with $pvec_1$ and $pvec_2$.}
\label{table:unbiased}
\end{table}
\section{Future Work}
\label{sec:discussion}

Our ground truth generation is based on object bounding box location, which is not accurate enough compared with instance-level segmentation. However, there is no available segmentation dataset labeled with object category information. As a future work, a dataset of instance-level segmented saliency map with categorical information about each object can be collected by labeling existing segmentation datasets.
Current model has non-differentiable operation layers, making the execution time depends on the object numbers in the input image. Revising the model to a differentiable one will be the major future work for us. The model can also be extended into focusing attention on the familiar person when combined with face recognition models. Another possible direction is extending current convolutional network into a recurrent version, thus it can find the personalized pattern about how the viewer changes the attention areas through a time series, which can be used for personalized scene description.


\section{Conclusion}
\label{sec:conclusion}
We design, implement, and evaluate a personalized attention prediction model PANet, which can predict saliency areas in an image according to individual user preference. This work presents the novel approach of including user preference in the vision model, allowing more efficient post-processing steps in various HCI applications. The shared feature extraction layers can efficiently provide multi-scale features for both saliency prediction and preference fitting stream. For training and validation, we collect saliency labels from two subjects together with their preferences. For new input preferences, we dynamically generate ground truth from established datasets, with parameters based on collected annotations. We experimentally validate that our saliency model can fit input preference and work for different preferences with a fast fine-tuning. Compared with general saliency prediction models, it fits much better to the saliency maps biased with personal tendencies. The more biased a preference is, the more performance gain provided by PANet.


\bibliographystyle{ACM-Reference-Format}
\clearpage
\bibliography{panet}

\end{document}